# Probabilistic Constraint Satisfaction with Non-Gaussian Noise

Russ B. Altman
Section on Medical Informatics
SUMC, MSOB X-215
Stanford, CA 94305-5479
altman@camis.stanford.edu

Cheng C. Chen
Dept. of Electrical Eng.
CIS, Rm 213
Stanford, CA 94305-4070
cchen@camis.stanford.edu

William B. Poland
Dept. of Eng.Econ.Systems
Terman Center 306
Stanford, CA 94305
poland@leland.stanford.edu

Jaswinder P. Singh
Dept. of Electrical Eng.
CIS, Rm 213
Stanford, CA 94305-4070
jps@samay.stanford.edu

## Abstract

We have previously reported a Bayesian algorithm for determining the coordinates of points in three-dimensional space from uncertain constraints. This method is useful in the determination of biological molecular structure. It is limited, however, by the requirement that the uncertainty in the constraints be normally distributed. In this paper, we present an extension of the original algorithm that allows constraint uncertainty to be represented as a mixture of Gaussians, and thereby allows arbitrary constraint distributions. We illustrate the performance of this algorithm on a problem drawn from the domain of molecular structure determination, in which a multicomponent constraint representation produces a much more accurate solution than the old single component mechanism. The new mechanism uses mixture distributions to decompose the problem into a set of independent problems with unimodal constraint uncertainty. The results of the unimodal subproblems are periodically recombined using Bayes' law, to avoid combinatorial explosion. The new algorithm is particularly suited for parallel implementation.

## 1 INTRODUCTION

Determining spatial coordinates from uncertain constraints is a problem that arises in many contexts, including the definition of biological molecular structure. Biological macromolecules (such as proteins or nucleic acids) contain hundreds to thousands of atoms, whose three-dimensional arrangements constitute their structure. The determination of molecular structure is critical for many pursuits in biomedicine, including the study of how molecules perform their function, and the design of drugs to augment or interfere with these functions. The primary sources of information about molecular structure are experimental, theoretical and empirical/statistical (Stryer 1991). However, of the 100,000 protein molecules that are estimated to be made within the human organism, the structures of only about 500 are known. The paucity of known structures derives, in part, from the great difficulty and expense of collecting experimental data of sufficient quantity and quality to allow each atom to be positioned accurately in three dimensions. In addition, theoretical and statistical constraints on structure (derived from biophysical models or from analysis of the previously determined structures, respectively) are also not sufficiently abundant or accurate to provide high resolution structural models by themselves. In combination, however, these data sources sometimes provide enough information to define the overall shape of a molecule or some elements of a high resolution structure. The focus of this work is to develop algorithms that are able to process uncertain data from multiple sources in order to produce an accurate model of a molecule.

Because the sources of data are uncertain (and in low abundance) the problem of defining structure is underdetermined. It is therefore necessary to estimate not merely a single structure that is consistent with the provided constraints, but also the variability in this structure. We have therefore developed an algorithm that is specifically geared towards providing estimates of structures as well as their uncertainty (Altman and Jardetzky 1989; Altman 1993). The algorithm represents a structure as a vector of mean coordinates, along with a variance/covariance matrix that summarizes the uncertainties in these coordinates. The random vector of mean values for Cartesian coordinates, $\mathbf{x}$, is of length $3N$ for $N$ atoms:

$$\mathbf{x} = \begin{bmatrix} x_1 & y_1 & z_1 & x_2 & y_2 & z_2 & \ldots & x_N & y_N & z_N \end{bmatrix}^{-1} \quad [1]$$

The vector of mean values of $\mathbf{x}$, $\hat{\mathbf{x}}$, is also of length $3N$. The variance/covariance matrix is of size $3N \times 3N$. The diagonal elements contain the variance of each of the elements of the mean vector. The off-diagonal elements contain the covariances between these elements:

$$\mathbf{C}(\mathbf{x}) = \begin{pmatrix} \sigma_{x_1x_1} & \sigma_{x_1y_1} & \cdot & \cdot & \sigma_{x_1z_N} \\ \cdot & \sigma_{y_1y_1} & & & \cdot \\ \cdot & & \cdot & & \cdot \\ \cdot & & & \cdot & \cdot \\ \sigma_{z_Nx_1} & \cdot & \cdot & \cdot & \sigma_{z_Nz_N} \end{pmatrix}$$

$$[2]$$

The mean vector and covariance matrix provide estimates of the positions, and summarize the the three-dimensional uncertainty of the atoms, as shown in Figure 1. The process of finding the optimal values for the parameters within the mean vector and covariance matrix is driven by external constraints on their values.

Constraints on structure have two components: a deterministic component, $h(x)$, that is a function of the coordinate vector, and an independent random component, $v$, (normally distributed, with mean of zero and variance of $C(v)$) describing the uncertainty in the value of the constraint, $z$:

$$z = h(x) + v \tag{3}$$

$$z \sim N\left(\mu_z, C(h(x)) + C(v)\right) \tag{4}$$

$$\mu_z = E[h(x)] \sim h(\hat{x}) + E(v) \sim h(\hat{x}) \tag{5}$$

When we say that constraint $z$ has a normal distribution, we are implying that it can be described by a Gaussian distribution with mean value taken from experimental, statistical or theoretical measurements, and variance $C(v)$ which is a property of the measurement technology. Thus, for example, one kind of constraint that is commonly used for determining molecular structure is the distance between atoms as measured by nuclear magnetic resonance (NMR) experiments. An NMR experiment might reveal that two atoms ($i$ and $j$) have a mean distance of 5 Ångstroms, with a variance of 2 $Å^2$. In this case, the function $h(x)$ is the scalar distance function, which depends on the three coordinates of the atoms $i$ and $j$. $v$ is the random variable that represents the error in the NMR measurement, normally distributed around 0 with variance of 2 $Å^2$. The mean value of $z$, $\mu_z$, is 5.

$$z = \sqrt{\left(x_i - x_j\right)^2 + \left(y_i - y_j\right)^2 + \left(z_i - z_j\right)^2} + v \tag{6}$$

Thus, given a model of the structure, which comprises the elements $\hat{x}$ and $C(x)$, we can compute an expected value for the distance and compare it with the measured value in the context of the expected noise, $v$, to see if they are compatible. If they are compatible, then we gain incremental confidence in our model, and the variances in $C(x)$ are reduced. If they are incompatible, then we make an appropriate update to our model ($\hat{x}$ and $C(x)$) to reflect the new information.

In previous work, we have shown that the model update can proceed in a Bayesian fashion, based on a modification of the extended, iterated Kalman filter (Gelb 1984) with measurement updates but no time updates. To summarize, a random starting $\hat{x}$ and $C(x)$ are created, with variances that are large (and consistent with the overall expected size of the molecule). Covariances are set to zero. Constraints are introduced serially and used to update both $\hat{x}$ and $C(x)$. The update equations are given by:

$$\hat{X}_{new} = \hat{X}_{old} + K[z - h(\hat{x}_{old})] \tag{7}$$

$$C(x)_{new} = C(x)_{old} - KHC(x)_{old} \quad \text{where} \tag{8}$$

$$K = C(x)_{old} H^T [HC(x)_{old} H^T + C(v)]^{-1} \quad \text{and} \tag{9}$$

$$H = \left.\frac{\partial h(x)}{\partial x}\right|_{\hat{x}} \tag{10}$$

If a set of uncertain distance constraints are used to update an estimate of $\hat{x}$ and $C(x)$, then the resulting new values of $\hat{x}$ and $C(x)$ will better "satisfy" the distance constraints (Altman 1993). *Satisfaction* is measured as the difference between the expected value of a constraint ($\mu_{z_i}$) and the observed value within the structural model ($h(\hat{x})$), divided by the standard deviation of the constraint noise:

$$E_i = \frac{\mu_z - h(\hat{x})}{\sqrt{C(v)}} \tag{11}$$

Because of inaccuracies introduced by the linearization of $h(x)$ shown in Equation 10, the simple serial introduction of the constraints does not converge to the best solution. However, if we use the new value of $\hat{x}$ after one round of introducing constraints as an improved starting point and repeat the procedure of introducing constraints iteratively, then we converge to a solution that satisfies all the constraints. We discuss the details of this iteration, and the similarity of our procedure, in some aspects, to simulated annealing in (Altman 1993). The resulting structural estimate provides both the mean value of the coordinates of each atom (in the $\hat{x}$ vector) and the uncertainty in these values (the diagonal of $C(x)$) and the covariation between these coordinates. This enables us to create structural illustrations such as shown in Figure 1 that demonstrate structure and level of uncertainty.

The procedure described above, assuming Gaussian constraint error, has been applied successfully to problems of analyzing uncertain experimental data (Arrowsmith, Pachter et al. 1991) and predicting structure from uncertain theoretical constraints (Altman 1993). The chief limitation has been that there are many sources of data that do not have normally distributed noise. As a result, the unimodal algorithm cannot adequately handle many cases of practical interest. For example, some types of theoretical constraints provide information that the distance between two points may be distributed in a trimodal manner.[1] A simple model of this constraint as a normal distribution is inadequate for capturing the information contained within the constraint. There are values that may appear likely in a Gaussian representation that actually fall between modes and are not likely. We have extended the algorithm to relax the assumption that all constraints have unimodal noise. The key insight is that any constraint can be approximated by a mixture of Gaussian distributions, which allows each of the components (unmixed Gaussians) to be treated by the original unimodal algorithm. We determine the number of components, and the mean values, variances and weights of each component in the mixture distributions, by the algorithms described in (Poland and Shachter 1993).

---

[1] Such a constraint might arise when the distance is conditioned on information that would allow one of three Gaussians to be selected (but that is unknown), and so the three possible components must be combined into a trimodal marginal distribution.

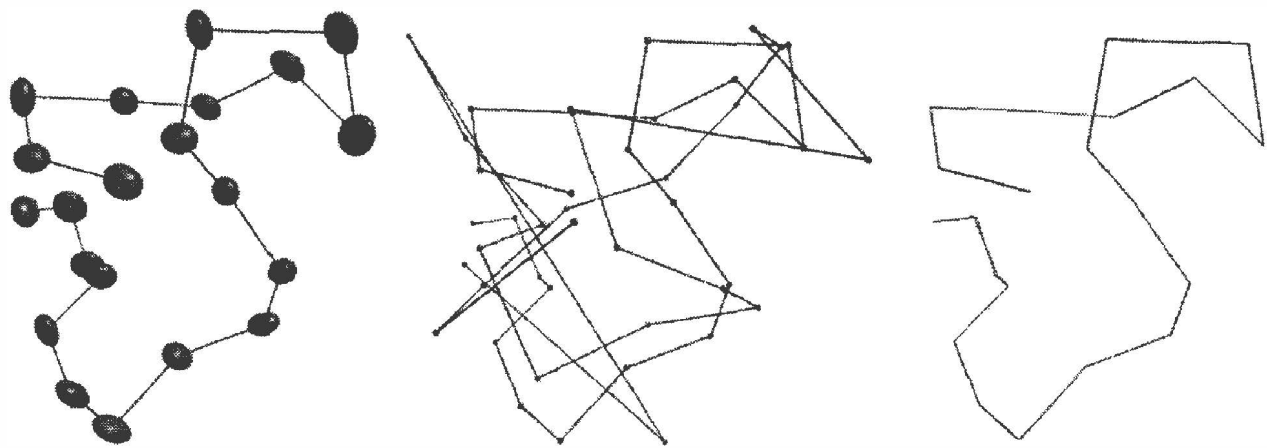

Figure 1. (LEFT) Typical output of the algorithm; We show here a fragment of a biological molecule with 21 atoms. Each atom has an uncertainty in three-dimensions described by an ellipsoid drawn at 2 standard deviations. Atomic bonds that connect two chemically adjacent atoms are drawn as lines (and are roughly 3.8 Å in length). Large ellipsoids indicate greater uncertainty in atomic location, and that atomic positions are less well defined by the constraints. The center of each ellipsoid is drawn at the mean position for each coordinate of the atom. The ellipsoid parameters are taken from the variance/covariance matrix which summarizes the uncertainties in the mean coordinates. (MIDDLE) Superposition of the known structure and the result produced by the unimodal algorithm in Experiment 1 (same orientation as shown on left). Although the structure produced by the unimodal algorithm has some general similarity to the gold standard structure, it has significant areas of mismatch. This mismatch illustrates the difficulties in reconstructing high resolution structures assuming a single component for all constraint noise. When detailed information about the distribution of constraint noise is available, these results suggest that it should be used. (RIGHT) Superposition of the known structure used for experiments in this paper, and the result produced by the multicomponent algorithm in Experiment 1. The two structures are identical to within 0.1 Å, and superimpose nearly perfectly.

Conceptually, we could generate all possible combinations of the constraint components and evaluate them with multiple runs of the unimodal algorithm. The particular set of components that best satisfies the constraints could then be identified. In fact, a distribution over these sets, each weighted by the degree to which they satisfy the constraints, could be produced. In practice, however, generating all possible component combinations is intractable. Instead, we generate combinations of components from a subset of constraints (shown in Figure 2B). We then solve each of these "partial" problems and recombine the results into a single, global, and improved estimate of $\hat{x}$ and $C(x)$. We then take the next group of constraints and repeat the process until all constraints have been introduced. As we have described previously for the unimodal algorithm, if the resulting structure estimate still has large errors, we can take the new estimate of $\hat{x}$ as a starting point for repeating the entire procedure. We reorder the constraints so that the least satisfied constraints are introduced first, and then repeat the cycle until we reach a stable estimate.

## 2  MULTICOMPONENT ALGORITHM

In order to understand the new algorithm, it is useful to view the unimodal algorithm graphically. As shown in

Figure 2A, we start with our initial estimate and derive improved estimates by introducing each constraint serially and updating our mean vector, $\hat{x}$, and covariance matrix, $C(x)$.[2]  Since each constraint is unimodal, we have no branching, and continue the iterative introduction of all constraints until the error metric converges to a stable value.

Our new algorithm can then be described graphically as shown in Figure 2B. Since each constraint has multiple components, we can imagine serially breaking each constraint into its constituent components and creating a number of parallel unimodal constraints. This produces an exponential fanning of the search space that becomes prohibitive. However, if we set a maximum depth D to which we are willing to fan, we can define a number of unimodal subproblems, each along a separate path of a tree of depth D, and solve each of these subproblems independently. Then, in order to reduce the combinatorics, we can recombine the solutions in order to, once again, have a single estimate of $\hat{x}$ and $C(x)$. By repeatedly fanning, solving the unimodal subproblems, and recombining, we can maintain a reasonable exponent, while not losing the advantage of the increased information content of multicomponent constraints. The only outstanding issue is: How do we combine the results of multiple unimodal problems using the available information about each of the constraint components (mean, variance, and weight)?

---

[2] In practice, we can actually introduce all the constraints simultaneously, or we can introduce them in groups. This decision is made based on the computational platform and the relative cost of operations such as matrix inversion. We have found that a group of 10 to 50 constraints at a time is optimal on many general purpose computers. In either case, we must iterate this process to overcome inaccuracies introduced in the linearization of $h(x)$.

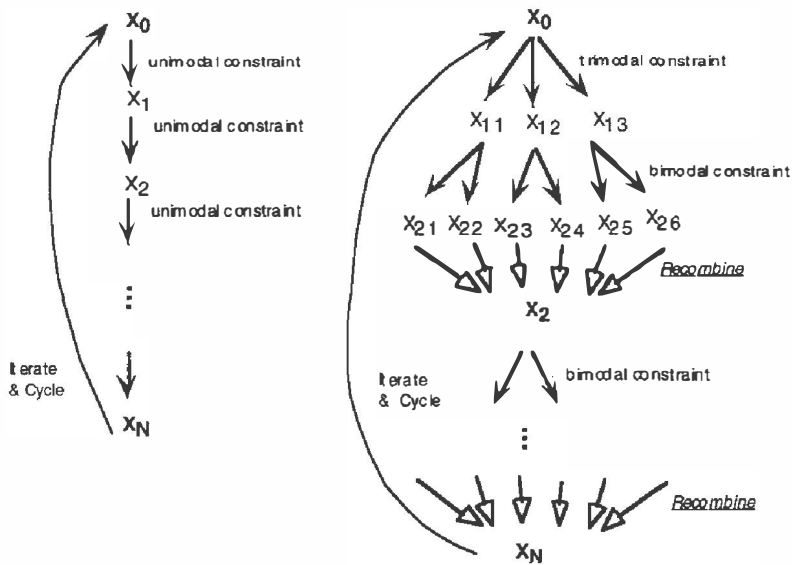

Figure 2. (A, LEFT) Strategy for the unimodal constraint algorithm. A starting estimate of the parameters (the vector, $x_0$) is serially modified by the introduction of constraints. If the residual errors are large, then the result is used as a starting point for another round of updating. (B, RIGHT) With the multicomponent algorithm, each constraint is described as a mixture of Gaussian distributions. The first constraint has three components, the second has two components, and these are combined to produce six branches. Branching continues until all resources are allocated. The results of individual calculations are recombined to calculate an intermediate mean vector and variance/covariance matrix The process is repeated with the next set of constraints. After many recombinations, a final estimate is produced, $x_N$, that can be used as a starting point if residuals are too high.

## 2.1   COMBINING RESULTS

For simplicity, we take the case of a scalar random variable $x$ subject to a multicomponent constraint $z$. Suppose *a priori* we have knowledge of the distribution of $x$:

$$x \sim N(\mu_X, \sigma_X^2) \qquad [12]$$

Constraint $z$ is described as a mixture of Gaussians:

$$z \sim a_1 N(\mu_{Z_1}, \sigma_{Z_1}^2) + \ldots + a_n N(\mu_{Z_n}, \sigma_{Z_n}^2) \qquad [13]$$

$$f(z) = a_1 f_{z_1}(z) + a_2 f_{z_2}(z) + \ldots a_n f_{z_n}(z) \qquad [14]$$

where the $z_i$'s are the components of the mixture random variable $z$ and the $a_i$'s are the prior weights associated with the component densities; that is, *a priori*, $z$ is equal to $z_i$ with probability $a_i$.

The root of the tree shown in Figure 2B is our prior knowledge (model) about $x$. The branches represent the possible outcomes of the mixture random variable $z$, weighted by the prior probability of the corresponding constraint component. Because we do not know which of the possible components $z_1, z_2 \ldots z_n$ the random variable $z$ actually takes on in the solution, we need to consider the possibility of following each path. Down at the leaves, each $\hat{x}_i$ comes from updating the original $x$ by constraint component $z_i$. To keep the amount of information manageable, we would like to find the posterior probabilities $w_1, w_2, \ldots w_n$ of the branches so that our updated knowledge about the value of $\hat{x}$ is represented as a weighted combination of the Gaussian $\hat{x}_i$

$$\hat{x} = \sum_{i=1}^{n} w_i \hat{x}_i \qquad [15]$$

The posterior weight $w_i$ of branch $i$ is P(branch $i$ | knowledge about $x$), which from Bayes' rule is:

$$w_i = \left[ \frac{P(\text{branch } i)\ P(\text{knowledge about x | branch } i)}{\sum_{j=1}^{n} P(\text{branch } j)\ P(\text{knowledge about x | branch } j)} \right] \qquad [16]$$

The denominator is simply a scale factor so that the probabilities sum to 1. The probability of branch $i$ is simply the component weight, $a_i$. The probability of the prior distribution on $x$, given the branch $i$, is a measure of how well the prior distribution of $x$ fits the distribution of constraint component $z_i$. This is related to the relative entropy of the two distributions and given by:

$$P(\text{knowledge about x | branch } i) = \left[ f_X(\mu_{z_i}) \exp\left( -\frac{1}{2} \frac{\sigma_{z_i}^2}{\sigma_X^2} \right) \right] \qquad [17]$$

The weight for each branch of the tree, therefore, is given by:

$$w_i = \left[ a_i f_X(\mu_{z_i}) \exp\left( -\frac{1}{2} \frac{\sigma_{z_i}^2}{\sigma_X^2} \right) \right] \qquad [18]$$

($w_i$ normalized to sum to 1). We are therefore able to calculate the new value of $\hat{x}$ based on the weights of each of the branches in the tree and the solution produced within that branch as shown in Eq. 12. In order to reduce the fan factor, we need to have the best Gaussian approximation to this distribution. The parameters of $x$ can be related to the parameters of the individual branch solutions by matching means and second moments:

$$\mu_{\hat{x}} = \sum_{i=1}^{n} w_i \mu_{\hat{x}_i} \qquad [19]$$

$$(\sigma_{\hat{x}}^2 + \mu_{\hat{x}}^2) = \sum_{i=1}^{n} w_i \left( \sigma_{\hat{x}_i}^2 + \mu_{\hat{x}_i}^2 \right) \qquad [20]$$

Using this machinery allows us to update our belief about the probability of each constraint component in the mixture based on the prior knowledge about the structure. With a randomly generated starting structure, the calculated posterior path weights may be so far from the solution as to be useless. We need to first get a rough estimate of the structure. One way is to collapse all the multicomponent constraints into representative unimodal constraints and run the unimodal algorithm for a few iterations with these constraints.

In order to evaluate a mean vector, $\hat{x}$, in the context of multicomponent constraints, we can calculate the distance of $h(\hat{x})$ from the mean of the nearest component in units of standard deviations (SD) and take the minimum distance to a component as the error for that constraint (analogous to Equation 3 above). Thus, the error for constraint $j$ with $m$ components each with mean value $z_i$ and standard deviation $C(v_i)$ is:

$$E_j = \min_{i=1,m} \left\{ \frac{z_i - h(\hat{x})}{C(v_i)} \right\} \qquad [21]$$

We have implemented this "branch and recombine" strategy in a program and describe some initial tests in the next section.

## 3.  EXPERIMENTS AND RESULTS

The long term goal of this work is to have an algorithm that converges to a correct solution (in terms of both mean positions for points, as well as their three-dimensional variances) given realistic biological data sets. Such data sets would contain constraints on distances, angles and perhaps other parameters of the structure. Distance constraints are usually the primary type of biological data that is available for estimating structure, and often they can not be simply summarized as a Gaussian distribution around some mean.  It is therefore useful to first evaluate the performance of the algorithm in the case of distance constraints with multiple components, even though it is designed to deal with any constraint that can be represented as a function of the atomic coordinates. For these experiments, we have taken a known structure of 21 atoms (a fragment of the molecule crambin (Hendrickson and Teeter 1981)) and generated the full set of exact distances between all atoms. We have then added various levels of spurious "noise" components in order to show that:

1.  Given a set of multicomponent distance constraints, the algorithm converges to the correct structure.

2.  The algorithm converges tolerates noise components at least to a level at which the correct components receive maximum weight (among the other components) only 50% of the time.

In general, there are $(N^2 - N)/2$ distance constraints between N atoms. However, only $4N-10$ exact distances are required to uniquely specify all $N$ positions. Thus, the $(N^2 - N)/2$ constraints actually overspecify the problem. In our test case, the 21 atoms have 210 total distances. For each calculation described below, we randomly generated starting coordinates (components of the x vector) in the range of 0 to 100 Å. The starting variances (in matrix $\mathbf{C(x)}$) were set to 100 Å$^2$ which is consistent with the overall expected size of a molecule with 21 atoms, such as we have chosen. The covariances in $\mathbf{C(x)}$ were all set to zero initially.

Experiment 1: For each distance between each pair of points, we created a synthetic multicomponent constraint. The real component (with mean taken from the known structure, and variance of 0.1) was given a weight between 0.5 and 1.0. A random number of "noise" components (ranging from 0 to 3) were then generated with means chosen randomly from 0.0 to 50.0 and variances chosen randomly from 0.0 to 10.0. These components were given equal weight, by equally dividing the remaining weight (that is, remaining after the assignment of weight to the real component between 0.5 and 1.0). We were then left with a set of multicomponent constraints with between 1 and 4 components, but which always had the predominant weight assigned to the actual component. Each of these multicomponent distributions was then collapsed into an equivalent normal (by taking a weighted average of the means and variances as described in Equations 19 and 20). This provided a set of single components that could be run through the old algorithm as a control.

The initial average error of the random structure(as calculated with Equation 9) for the unimodal constraints was 41.6 SD with a maximum error of 286 SD. The unimodal algorithm ran for 20 cycles and achieved a best average error for the constraints of 2.0 SD, with a maximum error of 12.4 SD.    This performance is consistent with that demonstrated previously for noisy constraints (Altman 1993).    The multicomponent algorithm was given the best solution produced by the unimodal algorithm and ran for 11 additional cycles, and achieved an average error of 0.07 SD with a maximum of 0.86 SD.    Figure 1B shows the known structure superimposed with the best solution produced by the unimodal algorithm. In contrast, Figure 1C shows the known structure from which constraints were created superimposed with the solution produced by the multicomponent algorithm.  They match to a root mean squared distance (RMSD) of 0.09 Å.  The RMSD between these two structures is 14.9 Å.  Figures 3A and

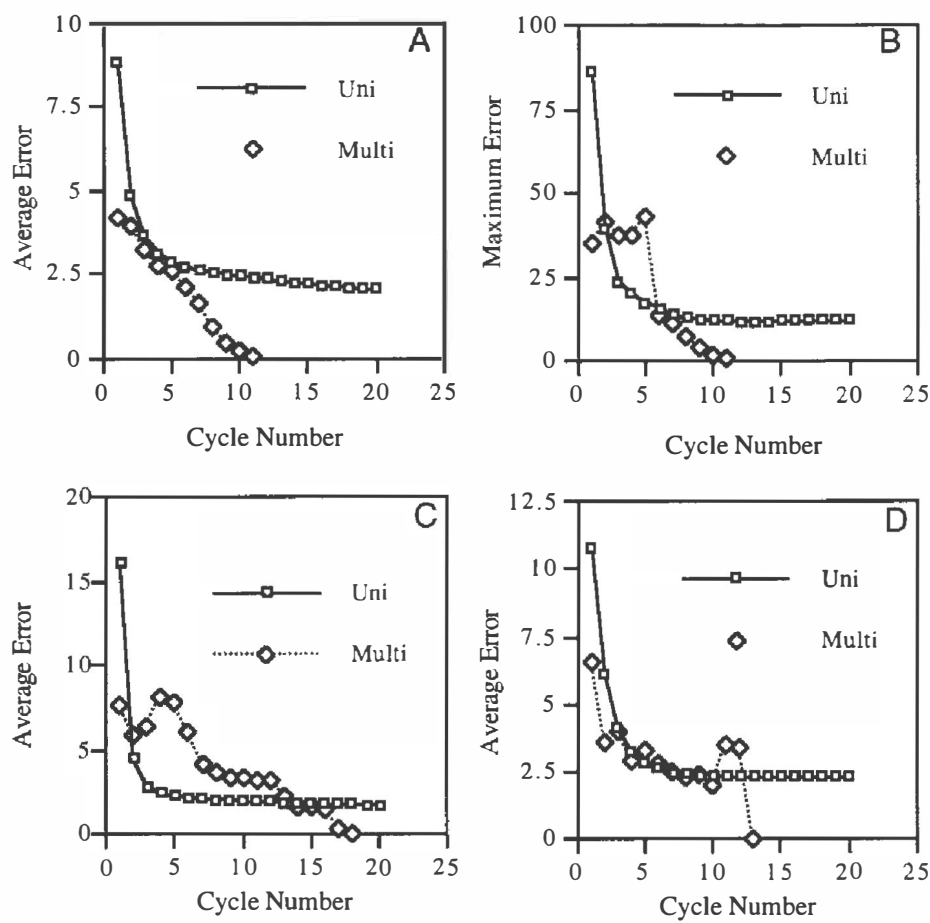

Figure 3. (A) Convergence of error as function of cycle number for Experiment 1. The unimodal algorithm plateaus at and average of 2.5 standard deviations (SD) for all constraints. Using this result as a starting point for the multicomponent constraint, allows it to converge to essentially zero error. (Note: the initial error for the multicomponent constraints is 4 SD and not 2.5 SD—as might initially be expected. The unimodal error is calculated based on a single, broad constraint error. The multicomponent error, as described in Eq. 21, is the distance of the measured value from the mean of the closest component. With the introduction of multiple components for each constraint, the variances of these components is much smaller than the single, broad component used for the unimodal calculation, and therefore the average distance from these components (the error) jumps to 4 SD.)

(B) Convergence of maximum error as a function of cycle number for Experiment 1.    At cycle 4, the algorithm jumps out of a local minimum in order to facilitate convergence to a globally lower error.  (C) Convergence of the average error in Experiment 2A as a function of cycle number.  Once again, the unimodal algorithm plateaus at 2 SD, but the multicomponent algorithm is able to find a much better structure, after exiting two local minima.  (D) Convergence of the average error in Experiment 2B as a function of cycle number.  Even for this problem, with more spurious noise introduced, the convergence of the multicomponent algorithm from the unimodal starting point is swift.

3B show the convergence rate of each of the two algorithms on this data set.    They show that the unimodal constraint representation plateaus at an average constraint error of 2 SD, while the multicomponent representation is able to take advantage of the increased constraint distribution precision to converge to an average error of less than 0.1 SD.

Experiment 2:  In order to test the ability of the algorithm to detect the correct solution as the weights of the spurious components increased, we generated two additional multicomponent constraint sets (along with the equivalent unimodal constraints as described for the previous experiment):

A. We set the weight of the real component randomly between 0.3 and 1.0 (instead of between 0.5 and 1.0 as in Experiment 1).  Once again, we generated between 0 and 3 noise components with means again chosen randomly from 0.0 to 50.0 and variances chosen from 0.0 to 10.0. The remaining weight was again distributed evenly. Now, however, approximately 30% of the constraints did not have a majority of the weight on the actual component.   Once again, the unimodal algorithm was run on a random starting structure, and reached a plateau at an average error of 1.8 SD and maximum error of 12.3 SD.  This solution was 15.7 Å RMSD from the known structure.    Using this as a starting point, the multicomponent algorithm converged to an average error of 0.003 (maximum error 0.03).  The resulting structure matched the gold standard to an RMSD of 0.002 Å. Figure 3C shows the convergence of the two algorithms as a function of cycle number.

B. We set the weight of the real component between 0.1 and 1.0.  We generated constraints analogously to the previous experiments.  Now, approximately half of the constraints did *not* weight the component describing the actual distance most highly. The unimodal algorithm produced a solution with average error of 2.3 SD (maximum of 17 SD). The solution was 23 Å RMSD from the actual structure. The multicomponent algorithm produced a structure with average error of 0.06 SD (maximum of 0.4 SD) that was 0.03 Å from the gold standard (Figure 3D).

## 4  DISCUSSION

The results of the first experiment demonstrate that the algorithm can recognize the correct components when offered a choice between the actual component from a true structure and randomly generated "noise" components. It also shows that a unimodal approximation to these non-Gaussian constraints does not contain information sufficient to converge to the correct structure. This observation is critical and supports our hypothesis that the unimodal assumption of our original algorithm was limiting its performance. In practice, we often have biological constraints that are known to have two or three sources of noise in addition to the signal. A Gaussian approximation allows us to get close (Figure 7 shows that the general topology of the unimodal solution is similar to the actual solution), but loses a large amount of information. It does provide, however, a useful starting point for further refinement. The recombination apparatus that are described in Equations 16-20 uses a Bayesian formulation to weight the structures that are produced by each of the branches. This apparatus relies on having a reasonable initial model of the structure with which to update. It appears from both the first and second experiments that the unimodal algorithm provides such a starting point, which allows the multicomponent algorithm to converge to the exact solution. Further experimentation is needed to more accurately characterize the radius of convergence of the multicomponent algorithm alone (that is, without the benefit of the unimodal solution).

The second experiment demonstrates that the "answer" component need not be the most highly weighted component in the constraint distribution in order for it to be recognized by the multicomponent algorithm. We successively reduced the average weight given to the actual component, and found that the algorithm reliably identified the gold standard answer. These experiments are somewhat limited because as the weights on other components are increased, all signal from the original structure may be lost. Nevertheless, they demonstrate that at least for the case where spurious components make up 50% of the distribution, the algorithm can still converge to the exact solution.

Both of the experiments demonstrate a capacity of the multicomponent algorithm to leave local optima that has been more extensively documented in the unimodal algorithm. Figures 9, 10 and 11 each show an error curve that has a peak between two local minima (at cycle 2 for Figure 9, cycle 4 for Figure 10, and cycles 3, 5 and 11 for Figure 11). Each of the minima represent solutions that were satisfactory for a large number of constraints (that is, they fell well within one of the constraint components), but which still had a large average error. Through the iterative reheating strategy of the algorithm (Altman 1993)(using the solution after introducing all constraints as a starting point, and resetting the covariance matrix to its initial value), it appears that the multicomponent algorithm has the same local minima-avoiding behavior that has been demonstrated in the unimodal algorithm (Altman 1993). This is not surprising since the rationale for the iteration steps does not make any assumptions about the manner in which the improved estimates of structure are generated from iteration to iteration.

## 5  RELATED WORK

There is a large literature in the processing of distance constraints between points to produce accurate structures. The *distance geometry* algorithm is based on an eigenanalysis of matrices that can be formed with knowledge of pairwise distances (Havel, Kuntz et al. 1983). This algorithm, in the case of sufficient exact distances, provides a closed form solution to the problem of determining structure from distances alone.[3] It is limited because in the case of sparse data, heuristic methods must be used to find solutions. It has been shown that the space of possible solutions is not uniformly sampled even when running this algorithm multiple times from different starting points (Metzler, Hare et al. 1989). It differs from our approach in two ways: it is not designed to handle constraints other than distances, and there is no probabilistic component to the interpretation of constraints. All distributions are assumed to be uniform between some minimum and some maximum.

Bayesian parameter estimation is reviewed in (Gelb 1984). We describe a method in which the assumption of Gaussian noise is relaxed. Simulated annealing (van Laarhoven and Aarts 1987) also uses an iterative technique to exit local optima, and is similar in concept in that respect, but not in implementation. We assume that distributions can be represented as mixtures of Gaussians, based on the results reported in (Poland and Shachter 1993).

This algorithm, in its most general form, is a type of parameter optimization. It differs from standard optimizations in that the value of the parameters (in the $x$ vector) are tracked along with their uncertainty and their covariances (in the $C(x)$ matrix). For this reason, it may have greater robustness to local optima, although this remains unproven. Shachter has proposed a method for finding the most likely posterior modes for a random variable, given Gaussian constraints (Shachter, Eddy et al. 1990) which is similar to the unimodal algorithm, but does not employ a reheating strategy to exit local minima. Our work differs in that it focuses specifically on non-Gaussian constraints. The new multicomponent algorithm (as can be seen in Figures 3 and 4) immediately suggests a parallel implementation. We are actively investigating this possibility.

---

[3]Theoretically there may be a symmetric solution that also satisfies all constraints. Biological molecules have chiral centers that usually dictate a single solution.

## 6   CONCLUSIONS

The determination of the positions of atoms in a biological molecule can be considered a constraint satisfaction problem. The sources of data (from biochemical measurements, our knowledge of basic chemistry, and from theoretical constraints) are the constraints. The goal is to find the sets of positions for all atoms that satisfy these constraints. Since these constraints are often noisy, it is critical that we maintain information about the reliability of our parameter estimates. In addition, many of these constraints do not have normally distributed noise, but have distributions that may be multimodal. A multimodal constraint implies that the value of the constrained parameter can take on values in more than one "neighborhood." In order to process constraint distributions which are not Gaussian, we have designed an algorithm for representing a constraint as a mixture of Gaussians. We use a branching strategy that is exponential, but controlled by intermittent recombination of the solutions.

In this paper, we introduced the algorithm and tested two necessary conditions for its applicability to the problem of biological structure determination. Specifically, we can draw the following conclusions:

1. That the algorithm converges to correct solutions given an over specified problem with little noise.

2. That the algorithm converges to correct solutions even when the weights on spurious or "noise" components in the constraints are, on average, greater than 50% of the total weights.

3. That the multicomponent algorithm outperforms the unimodal algorithm for equivalent data sets, in part because of its ability to more finely represent constraint noise distributions.

4. That the output of the unimodal algorithm provides a good starting point for the multicomponent algorithm.

5. That the multicomponent algorithm seems to successfully avoid local minima, as has been demonstrated in the unimodal algorithm. In the latter case, this has been shown to be due to the strategy of iterative refinement of the parameter mean estimates, with periodic "reheating" of the variance estimates to allow unsatisfied constraints to make relatively large perturbations on the parameter estimates.

### Acknowledgments

RBA is a Culpeper Foundation Medical Scholar, and receives computer support from the CAMIS resource, NIH grant LM05305. Graphical display software provided by S. Ludtke. We acknowledge an equipment grant from Hewlett Packard.    JPS is supported under DARPA Contract N00039-91-C-0138, and thanks John Hennessy for his support.